\pdfoutput=1

\documentclass[11pt]{article}

\usepackage{coling}

\usepackage{times}
\usepackage{latexsym}
\usepackage{url}
\usepackage{array}
\usepackage{graphicx}
\usepackage{geometry}
\usepackage{float}
\usepackage{xspace}
\usepackage{booktabs}
\usepackage{tabularx}
\usepackage{tabularray}
\usepackage{arydshln}
\usepackage{longtable}

\usepackage[T1]{fontenc}

\usepackage[utf8]{inputenc}
\usepackage{CJKutf8}

\usepackage{microtype}

\title{Continual Pre-Training is (not) What You Need in Domain Adaption}

\author{
  Pin-Er Chen \and Da-Chen Lian \\
  \textbf{Shu-Kai Hsieh} \and \textbf{Sieh-Chuen Huang} \\  
  National Taiwan University \\
  Taipei, Taiwan
  \AND 
  Hsuan-Lei Shao \\
  Taipei Medical University \\
  Taipei, Taiwan 
  \AND 
  Jun-Wei Chiu \and Yang-Hsien Lin \and Zih-Ching Chen \\ 
  \textbf{Cheng-Kuang Lee} \and \textbf{Eddie TC Huang} \and \textbf{Simon See} \\ 
  NVIDIA AI Technology Center, NVIDIA Corporation \\
  Santa Clara, CA, USA
}

\newcommand{\modelname}[1]{\textsc{#1}}

\begin{document}
\begin{CJK*}{UTF8}{bsmi}
\maketitle
\begin{abstract}
The recent advances in Legal Large Language Models (LLMs) have transformed the landscape of legal research and practice by automating tasks, enhancing research precision, and supporting complex decision-making processes. However, effectively adapting LLMs to the legal domain remains challenging due to the complexity of legal reasoning, the need for precise interpretation of specialized language, and the potential for hallucinations. This paper examines the efficacy of Domain-Adaptive Continual Pre-Training (DACP) in improving the legal reasoning capabilities of LLMs. Through a series of experiments on legal reasoning tasks within the Taiwanese legal framework, we demonstrate that while DACP enhances domain-specific knowledge, it does not uniformly improve performance across all legal tasks. We discuss the trade-offs involved in DACP, particularly its impact on model generalization and performance in prompt-based tasks, and propose directions for future research to optimize domain adaptation strategies in legal AI.

\end{abstract}

\section{Introduction}

The advent of legal AI represents a significant transformation in the delivery of legal services and the execution of legal research. As AI technologies advance, they provide unprecedented capabilities for automating routine tasks, improving the precision of legal research, and supporting complex decision-making processes. Legal LLMs, in particular, can process vast quantities of legal texts, precedents, and statutes with a level of speed and accuracy beyond human capabilities, offering critical insights for legal analysis and strategy. Furthermore, these advancements have the potential to democratize access to legal services, making expert-level legal analysis more accessible to individuals and organizations that might otherwise lack the necessary resources~\citep{lai2023largelanguagemodelslaw}.

However, the integration of legal-domain knowledge into LLMs poses significant challenges, particularly in ensuring ethical use, maintaining transparency in decision-making, and addressing concerns about bias. Among these challenges, the most crucial is enhancing the reasoning capabilities of LLMs~\citep{Almeida_2024}. Legal reasoning is a complex process that involves interpreting statutes, case laws, and regulations, and applying them to specific \textit{facts}. Unlike other forms of logical reasoning, legal reasoning demands an understanding of the precise and normative meanings of legal language, which is often highly specialized and context-dependent~\citep{bongiovanni2018handbook}. 

Additionally, it requires grasping the underlying legislative principles and intentions. In logical terms, for the most part, legal reasoning represents a form of \textit{defeasible} reasoning/argumentation that is based on the interpretation of legal texts and norms. Such an approach must not only consider the semantic meaning of words and expressions, but also their pragmatic aspects in a communicative context, reasoning together to deliberate to carry out our collective goals and values. An effective legal AI must be capable of reasoning across these dimensions, offering interpretations and arguments that align with established legal standards.

Given these challenges, Domain-Adaptive Continual Pre-Training (DACP) offers a promising solution for improving LLMs' legal reasoning and reducing the occurrence of hallucinations. DACP enables LLMs to continuously adapt to the evolving legal landscape by scaling up new legal texts, cases, and regulations into the model's training data~\citep{colombo2024saullm54bsaullm141bscaling}. It is noteworthy that most existing research on DACP in the legal domain has been conducted within Anglo-American legal systems. This paper seeks to break new ground by focusing on the Taiwanese-Mandarin legal system, which has been adapted from and influenced by the Continental legal system.

The rest of the paper is organized as follows. We briefly review the literature on legal reasoning and related evaluation benchmarks. In Section \ref{sec:models}, we present our models, which have been trained on legal data\footnote{The Hugging Face repository will be available after the anonymized review}. Section \ref{sec:experiment} describes the construction of our benchmark, specifically designed to assess the legal reasoning capabilities of LLMs within the Taiwanese legal framework. The benchmark consists of multiple tasks: single-multiple choice questions (\ref{sec:single-multiple-choice}), argument-based decision-making in legal symposia (\ref{sec:legal-symposium}), and essay questions (\ref{sec:essay-question}). The experiment results of the LLMs are illustrated respectively. Section \ref{sec:conclusion}
 concludes the paper, and Section \ref{sec:limitations} discusses the limitations and future work.

\section{Related Work}

\textbf{Domain-Adaptive Continual Pre-Training.} Building on the success of English LLMs in general NLP tasks, several studies have investigated language-specific and task-specific continual pre-training methods \citep{cui2024textencoding, guo2023continuous, zhao2024llamaenglish, zheng2023marinegpt}. Domain-Adaptive Continual Pre-Training (DACP) refines domain-specific LLMs by persistently pre-training a general-purpose model on an extensive corpus of domain-specific unlabeled data \citep{guru2020dacp, jin2022cap, shi2024continual}. This method enables the model to more accurately align with the in-domain distribution, improving its adaptation to the specific demands of the domain \citep{wu2022continual, xie2023dacp, yildiz2024cp}. Examples include BioBERT \cite{lee2019biobert}, PubMedBERT \cite{gu2021pubmedbert}, BloombergGPT \cite{wu2023bloomberggpt}, and EcomGPT-CT \cite{ma2023ecomgpt}, which are tailored to the biomedical or financial sectors. While some research reveals outstanding performance using this pre-training technique, many models exhibit different results, such as general knowledge forgetting or a trade-off between domain-specific tasks and general NLP tasks like Named Entity Recognition (NER) or long-context understanding \citep{chen2020forget, ma2023ecomgpt, yang2024medical, zhang2023xuanyuan20, zheng2023marinegpt}.

Domain-Adaptive Continual Pre-Training (DACP) may not provide consistent benefits across all task types, as its effectiveness can vary depending on the specific nature of the tasks. \citet{cheng2023adapting} observed that while DACP enhances domain-specific knowledge within a language model, it unexpectedly reduces performance on prompting tasks across various domains. They propose that although this pre-training improves fine-tuning by enriching domain knowledge, it simultaneously diminishes the model's effectiveness in prompt-based tasks, suggesting a trade-off between these two aspects. In addition, \citet{xie2023dacp} conducted a comparative study between FinPythia and Pythia models, both with and without the application of DACP. Their findings reveal that while FinPythia demonstrates superior performance in tasks such as FPB, Headline, and NER compared to its Pythia counterparts, it underperforms in the FiQA Sentiment Analysis task. These imply that the advantages of DACP may not be evenly realized across different types of tasks.

\textbf{Evaluation Benchmarks} 
Recent advancements in LLMs have highlighted the need for refined benchmarks to accurately assess their capabilities across various domains. For language-specific contexts, benchmarks like C-Eval \cite{huang2023ceval}, CMMLU \cite{li2024cmmlu}, and TMMLU+ \cite{tmmlu+2024} have been developed. C-Eval tests Chinese language LLMs with multiple-choice questions across four difficulty levels, covering 52 subjects, including humanities, science, and engineering, with a subset focused on advanced reasoning. 
CMMLU and TMMLU+, modeled after MMLU~\cite{hendrycks2021measuringmassivemultitasklanguage}, evaluate LLMs in Simplified and Traditional Chinese, respectively, across subjects ranging from elementary to professional levels, integrating real-world and cultural knowledge. 
In the legal domain, specific benchmarks like LawBench \cite{fei2023lawbench}, LAiW~\cite{dai2024laiw}, and LegalBench \cite{guha2023legalbench} offer assessments of LLM performance in legal tasks, testing memory, understanding, application, and complex legal reasoning through tasks aligned with real-world scenarios and judicial standards. DISC-LawLLM-Eval \cite{yue2023disclawllm} separates tasks into objective evaluations with multiple-choice questions from Chinese judicial exams and subjective evaluations where LLMs are scored on question-answering accuracy by GPT-3.5 Turbo.

Current legal benchmarks primarily categorize tasks by legal reasoning complexity, with the most challenging tasks involving legal application and generation, though multiple-choice and classification tasks dominate. While complex tasks like case understanding and legal consultation are included, further research is needed to determine how effectively these benchmarks evaluate LLMs in legal reasoning tasks.

\section{Models and Methods}
\label{sec:models}

We train several models, either through pre-training followed by full-parameter fine-tuning or through Low-Rank Adaptation.\footnote{The models described in this paper will be made publicly available upon acceptance for publication.}

For \modelname{Llawa}, we use \modelname{Llama3-TAIDE-LX-8B-Chat-Alpha1}(\modelname{TAIDE})\footnote{\url{https://huggingface.co/taide/Llama3-TAIDE-LX-8B-Chat-Alpha1}} as our base model for continuously pre-training Llawa.
It was continuously pre-trained from Meta's \modelname{Llama-3-8B} \citep{meta_llama3_2024}\footnote{\url{https://huggingface.co/collections/meta-llama/meta-llama-3-66214712577ca38149ebb2b6}} on 43B tokens of Traditional Chinese that reflect the linguistic and cultural characteristics of Taiwan. 
We chose this model with the belief that a domain-adapted, culturally aware model would be beneficial to downstream tasks involving local law.
We carry out three stages of training:
\begin{enumerate}
    \item We perform Domain-Adaptive Pre-Training on legal documents from Taiwan. This step helps the base model learn knowledge relevant to Taiwan's legal system.
    \item We perform full-parameter instruction tuning on several law-related tasks. This teaches the model to answer legal questions helpfully.
    \item We perform preference alignment on the instruction-tuned model from the previous stage. Preference alignment often helps improve the model's output in becoming more helpful (e.g., more informative, less harmful)
\end{enumerate} 

Besides training \modelname{Llawa}, we instruction-tune two additional models using Low-Rank Adaptation (LoRA)~\citep{hu2021loralowrankadaptationlarge}:
\begin{enumerate}
    \item \textbf{\modelname{Bllawa}}, fine-tuned from \modelname{Meta-Llama-3-8B-Instruct}\footnote{\url{https://huggingface.co/meta-llama/Meta-Llama-3-8B-Instruct}}
    \item \textbf{\modelname{Blawstral}}, fine-tuned from \modelname{Mistral-Nemo-Instruct-2407}\footnote{\url{https://huggingface.co/mistralai/Mistral-Nemo-Instruct-2407}}
\end{enumerate}

The purpose of training these two models is to see if performing only LoRA can achieve comparable results. LoRA is a parameter-efficient method of fine-tuning a model and is more accessible to those with fewer resources. 
Furthermore, \modelname{TAIDE} was continuously pre-trained from \modelname{Meta-Llama-3-8B}. By fine-tuning \modelname{Llama-3-8B-Instruct} on our legal tasks, we can observe how the additional knowledge gained from \modelname{TAIDE}'s pre-training and our pre-training stages benefit the final model.

\subsection{Domain-Adaptive Pre-Training in the Legal Domain}

\begin{table}
\begin{tabularx}{0.46\textwidth}{X >{\raggedleft\arraybackslash}X >{\raggedleft\arraybackslash}X}
\toprule
\textbf{Dataset} & \textbf{Size (GB)} & \textbf{Tokens (B)} \\
\midrule
Taiwan Law             & 112.36             & 56                  \\
German Law             & 41.23              & 41                  \\
Self-Curated           & 10.37              & 10                  \\
\bottomrule
\end{tabularx}
\caption{Datasets used for pre-training. Token counts are calculated using the base model's tokenizer.}
\label{tab:pre-training-ds}
\end{table}

Information regarding the pre-training datasets can be found in Table~\ref{tab:pre-training-ds}. 
\textit{Taiwan Law} contains publicly available data from Judicial Yuan,\footnote{\url{https://www.judicial.gov.tw/tw/mp-1.html}} including laws and regulations, as well as court documents. 
\textit{German Law} is a subset of the MultiLegalPile\footnote{\url{https://huggingface.co/datasets/joelniklaus/Multi_Legal_Pile_Commercial}}, which is a multilingual legal dataset containing case law, contracts, legislation, and others. We use the \textbf{de} subset that contains all German language data. This was added because of the influence Germany's legal system has had on Taiwan's legal system \cite{ZhangTaiwan}.

\textit{Self-Curated} contains data such as knowledge graphs from ConceptNet\footnote{\url{https://huggingface.co/datasets/conceptnet5/conceptnet5}} for English, German, and French and the Chinese Buddhist Electronic Texts Association\footnote{\url{https://www.cbeta.org/}} (CBETA). The considerations for selecting these data include enhancing logical reasoning training, providing reference knowledge for civil law systems, and familiarizing students with Classical Chinese (since the legal language style in Chinese closely resembles Classical Chinese).

\subsection{Continual Pre-Training}
Pre-training for \modelname{Llawa} was done on 2 x NVIDIA DGX H100 nodes (16 H100 GPUs). These resources were accessed via the Taipei-1 supercomputer located in the Kaohsiung Software Park. We used NVIDIA NeMo\footnote{\url{https://github.com/NVIDIA/NeMo}} and NVIDIA NeMo Framework Launcher\footnote{\url{https://github.com/NVIDIA/NeMo-Framework-Launcher}} to run our training scripts. We used FP8 mixed-precision training to take advantage of the memory savings and faster training afforded by the hardware. We use a tensor parallel size of 2.
We trained for one epoch, using a global batch size of 224 and a sequence length of 8192. Training took approximately 8 days. We start with a learning rate of 1e-4 and gradually decrease it using a cosine decay schedule.

\subsection{Instruction Tuning}

Instruction tuning serves to bridge the gap between pre-training's next token prediction task on unlabeled data and the task of being a helpful model to users. This entails training the model on (\texttt{INSTRUCTION}, \texttt{OUTPUT}) pairs \citep{zhang2024instructiontuninglargelanguage}.

\subsubsection{\modelname{Llawa} Training Details}
Full-parameter instruction tuning was performed on 4 x NVIDIA RTX 6000 Ada using Axolotl \citep{axolotl}. 

We train two instruction-tuned versions of Llawa.
The first version, \modelname{Llawa-TC-YZL-Instruct}, was trained in two stages. For the first stage, the base \modelname{Llawa} model was trained on a cleaned version of \modelname{TaiwanChat}~\citep{taiwanllm}.\footnote{\url{https://huggingface.co/datasets/yentinglin/TaiwanChat}} 
This stage is meant to allow the model to learn general instruction following capabilities.
\modelname{TaiwanChat} is an instruction dataset comprising other instruction datasets, which were translated to traditional Chinese using \modelname{GPT-3.5-Turbo}. 
We use a cleaned version that removed duplicates and malformed conversations (e.g., conversations ending with the user instead of an assistant's response). 
We trained on this dataset for three epochs. 
The second stage further trains the model on legal-related tasks, the types of which can be found in Table~\ref{tab:task-description}. 
This stage can be seen as specializing the model to answer more legal-oriented questions. We train for two epochs, which is when validation loss fails to improve.

The second version, \modelname{Llawa-TCxYZL-Instruct}, was fine-tuned in one stage by combining \modelname{TaiwanChat} and our legal dataset and shuffling the resulting dataset. Training on both the general instruction dataset and the legal instruction dataset simultaneously can reduce the likelihood of the model forgetting any generalizable skills that would be beneficial for completing any tasks. We trained for two epochs, which is when validation loss did not continue to improve.  

\subsubsection{\modelname{Bllawa} \& \modelname{Blawstral} Training Details}
\modelname{Bllawa} and \modelname{Blawstral} were trained using Low-Rank Adaptation (LoRA) on 1 x NVIDIA A100 80GB. We use the Unsloth \citep{unsloth} library, which uses custom-written kernels that reduce memory usage and training time. 
We add adapter weights to all linear layers. We use a rank of 64, an alpha of 128 with a learning rate of 5e-5. We train for three epochs on our legal dataset. 

\subsection{Preference Alignment}
Preference alignment is often included as a post-training step. The purpose of this stage is to have the model learn desirable responses and behaviors from the knowledge it obtained in previous training steps. For example, while a coding model should understand common programming mistakes, it should output high-quality, correct code (unless instructed otherwise)~ \citep{rafailov2024directpreferenceoptimizationlanguage}.

We experiment with two preference alignment methods: Direct Preference Optimization (DPO) and Odds Ratio Preference Optimization (ORPO). DPO is often used because of its simplicity compared to other methods, such as reinforcement learning from human feedback (RLHF), by optimizing a simple binary cross entropy objective between preferred and dispreferred responses \citep{rafailov2024directpreferenceoptimizationlanguage}.

While DPO requires instruction-tuning beforehand, ORPO performs model alignment during the supervised fine-tuning stage by including an odds ratio-based penalty to the negative log-likelihood loss (NLL), thus simplifying post-training further~ \citep{hong2024orpomonolithicpreferenceoptimization}. 
While ORPO is usually used from a base model before instruction-tuning, we wish to keep the number of potential confounding factors to a minimum (i.e., the order and number of post-training steps) and so perform preference alignment with ORPO after instruction-tuning.
Preference optimization was performed using 2 x NVIDIA A100 80GB GPUs. We use TRL's~\citep{vonwerra2022trl} implementation of DPO and ORPO. We set DPO's and ORPO's \texttt{beta} parameters to \texttt{0.01} and \texttt{0.1}, respectively. We train for a maximum of three epochs or until loss on the evaluation dataset converges.

\begin{table*}[h!]
\small
\centering
\begin{tabular*}{0.9\textwidth}{>{\centering\arraybackslash}p{3em}>{\raggedright\arraybackslash}m{0.22\textwidth}>{\centering\arraybackslash}p{4em}>{\raggedright\arraybackslash}m{0.22\textwidth}>{\raggedright\arraybackslash}m{0.22\textwidth}} 
\toprule
\textbf{Task} & \textbf{Data Source} & \textbf{Instances} & \textbf{Input} & \textbf{Output} \\
\hline
A & Bar and Judicial Exam & 662 & Multiple-choice question with 4 options & Answer (\textit{sg.}) \\
\midrule
B & Taiwan Jurist Journal & 242 & Multiple-choice question with 4-6 options & Answer (\textit{sg./pl.})\\
\midrule
C & Legal Symposium of Taiwan High Court & 1854 & Issue for discussion with multiple arguments & Final argument \\
\midrule
D & Bar and Judicial Exam & 40 & Hypothetical legal scenario & Essay \\
\bottomrule
\end{tabular*}
\caption{Overview of Tasks A, B, C, and D. \textit{sg.} represents answer with single option, while \textit{pl.} represents an answer with multiple options.}
\label{tab:task-description}
\end{table*}

\begin{table*}[ht]
\centering
\resizebox{0.9\textwidth}{!}{%
\tiny
\begin{tabular}{lcccc}
\toprule
\textbf{Model} & \textbf{Task A} & \textbf{Task B} & \textbf{Task C} & \textbf{Task D} \\
\midrule
gpt-4-turbo & 53.93 & 43.80 & 49.43 & 80.85 \\
\hdashline
Mistral-Nemo-Instruct-2407 & 27.19 & 30.99 & 51.28 & \textbf{74.26} \\
Blawstral & \textbf{37.76} & \textbf{37.19} & \textbf{56.54} & 62.03 \\
\hdashline
Meta-Llama-3-8B-Instruct & \textbf{38.82} & \textbf{38.84} & 51.64 & \textbf{68.25} \\
Bllawa & 36.56 & 32.64 & \textbf{52.18} & 51.42 \\
\hdashline
Llama-3-TAIDE-LX-8B-Chat-Alpha1 & 24.92 & \textbf{33.47} & 51.40 & 77.62 \\
Llawa-TC-YZL-Instruct & 25.38 & 30.17 & \textbf{53.49} & - \\
Llawa-TCxYZL-Instruct & \textbf{28.55} & \textbf{33.47} & 53.07 & - \\
Llawa-TCxYZL-ORPO & 25.38 & 29.34 & 45.67 & - \\
Llawa-TCxYZL-DPO & 24.47 & 28.51 & 43.94 & - \\
\bottomrule
\end{tabular}%
}
\caption{Model performance of the three tasks. The metric is accuracy (\%) for Tasks A, B, and C. Task D used GPT-4o as the evaluator in which each model's response was graded against the reference answer.}
\label{tab:model-performance}
\end{table*}

\section{Experiments}
\label{sec:experiment}

We create four legal reasoning tasks as datasets,~\footnote{The data for Tasks A, B, and D were sourced from Taiwan's Bar and Judicial Exam, while the data for Task C were gathered from the Taiwan High Court website (\url{https://tph.judicial.gov.tw/tw/np-1131-051.html}).} detailed in Table~\ref{tab:task-description}.
We conduct a comparison between our models and other open-source models. For each model, we employ greedy decoding for generation. The maximum input token length is 7192, with prompts exceeding this limit truncated on the right. All models are assessed in one-shot settings, with an example question-answer pair following the instruction (refer to the Appendix for task-specific prompts).

\subsection{Multiple Choice Questions}
\label{sec:single-multiple-choice}
Table~\ref{tab:model-performance} demonstrates the model performance on all tasks. For tasks A and B, it is not surprising that \modelname{gpt-4-turbo} still outperforms the others on this type of multiple-choice question-answering task. 
While fine-tuning is generally linked to improved performance on specific tasks, our findings suggest a more complex situation. Interestingly, \modelname{Meta-Llama-3-8B-Instruct} outperforms \modelname{Bllawa} on Tasks A and B despite \modelname{Bllawa} being fine-tuned for these specific tasks. 
In contrast, the expected results were observed with \modelname{Blawstral}, derived from \modelname{Mistral-Nemo-Instruct-2407}, where fine-tuning led to significant performance improvement. 

The decline in performance from \modelname{Meta-Llama-3-8B-Instruct} to \modelname{Bllawa} may be due to the former's extensive post-training processes, including supervised fine-tuning, rejection sampling, and DPO. This thorough training likely produced a versatile model with strong reasoning and general test-taking abilities. 
We postulate that our fine-tuning process may have unintentionally pushed \modelname{Bllawa} into a suboptimal state, possibly improving in-domain text generation at the cost of essential test-taking skills needed for our evaluation tasks.

This phenomenon raises an important research question about the optimal balance and amount of training data required to develop a base model that retains critical skills while also excelling in targeted tasks. Future investigations into this balance could provide valuable insights for advancing language model development and fine-tuning strategies.

As for \modelname{Llawa-TCxYZL-Instruct}, derived from \modelname{Llama-3-TAIDE-LX-8B-Chat-Alpha1}, it shows better performance on Task A and similar performance on Task B. Since \modelname{Llawa-TCxYZL-Instruct} performs only slightly better than its base model (\modelname{Llama-3-TAIDE-LX-8B-Chat-Alpha1}), we investigated the efficacy of preference optimization techniques, specifically DPO and ORPO, often demonstrated to enhance model performance in prior art. 
Our methodology involved designating the ground truth as the preferred output and the model's generated output as the rejected one. Contrary to expectations, these techniques did not yield performance improvements in our targeted tasks.

Several hypotheses may account for this unexpected outcome:
(1) The designation of the ground truth as the preferred answer and the model output as the rejected answer may not be optimal for this specific training paradigm. 
(2) Limited variability in the training data may have led to rapid overfitting on dataset noise, persisting even when controlling for the number of training epochs. 
(3) The paucity of literature on preference optimization best practices may have resulted in suboptimal hyperparameter selection given the specific circumstances of our study. 
Due to time and budget constraints, a comprehensive exploration of potential training parameters was not feasible within the scope of this study. 
We propose this as a promising avenue for future research, which could significantly contribute to our understanding of preference optimization techniques in language model fine-tuning and potentially resolve the performance discrepancies observed in our current work.

\begin{table*}[!h]
\centering
\resizebox{\textwidth}{!}{
\begin{tabular}{p{0.4\textwidth} p{0.4\textwidth} p{0.4\textwidth} p{0.4\textwidth}}
\toprule
\multicolumn{2}{l}{\textbf{Input}} & \multicolumn{2}{l}{\textbf{Output}}  \\
\midrule
Issue & \multicolumn{1}{l}{Opinions} & Final Argument & \multicolumn{1}{l}{Review/Research Opinion} \\
\midrule
\multicolumn{1}{p{0.4\textwidth}}{After the commencement of a juvenile court trial, can the victim of a crime in a juvenile case file a supplementary civil lawsuit to seek damages?} & \multicolumn{1}{p{0.4\textwidth}}{Opinion A: According to Article 1 of the Juvenile Delinquency Act, which states... Opinion B: The nature of juvenile corrective measures is fundamentally different...} & \multicolumn{1}{p{0.4\textwidth}}{Opinion B is adopted. The Juvenile Delinquency Act explicitly enumerates the provisions of the Code of Criminal Procedure that apply (such as in Articles 16 and 24 of the Juvenile Delinquency Act). Since there is no provision in the Juvenile Delinquency Act for applying the supplementary civil lawsuit procedures of the Code of Criminal Procedure, they cannot be applied.} & \multicolumn{1}{p{0.4\textwidth}}{Judicial Yuan Second Division Research Opinion: The research opinion agrees with the discussion result and finds Opinion B to be correct. However, the issue originally referred to "the crime victim in a juvenile case," which is a misnomer and should be corrected to "the victim in a juvenile corrective measure case." As for the crime victim in a juvenile criminal case, they may still file a supplementary civil lawsuit.} \\
\bottomrule
\end{tabular}
}
\caption{An instance of legal symposium. For brevity, the detailed statements of the two opinions are omitted. The full version in Taiwan Mandarin will be presented in the Appendix.}
\label{tab:legal-symposium-example}
\end{table*}

\subsection{Argumentation in Legal Symposium}
\label{sec:legal-symposium}

For Task C, we prompt the LLMs to select a specific stance based on the arguments within a legal symposium. It is noted that the generated answer is not merely a classification; the possible answers can vary (e.g., adopt part or whole argument from A/B/C, agree with the review opinion, affirm/deny a certain argument, reserve judgment, refer the case to the High Court, etc.), depending on the context, characteristics of the legal case, and the number of arguments. 

As indicated in column Task C of Table~\ref{tab:model-performance}, \modelname{Blawstral} achieves the highest ranking, followed by \modelname{Llawa-TC-YZL-Instruct} and \modelname{Llawa-TCxYZL-Instruct}. Compared to previous tasks, \modelname{Bllawa} outperforms its base model, \modelname{Meta-Llama-3-8B-Instruct}, in this task. This underscores the models' effectiveness in generating final opinions from legal arguments. 
Notably, \modelname{gpt-4-turbo}, \modelname{Mistral-Nemo-Instruct-2407}, and \modelname{Meta-Llama-3-8B-Instruct} demonstrate comparatively lower accuracy on this task. 
The results may be influenced not only by the volume of training data related to the Taiwanese legal system but also by the unique nature of the legal symposium setting. In such a setting, factual knowledge used in the arguments might be incomplete or contradictory, with both sides presenting imperfect arguments to reach a provisional conclusion based on the relevant standard of proof for the case \citep{bongiovanni2018handbook}. While the LLMs are prompted to formulate a closing statement, they require both general logical reasoning and legal knowledge.

The final arguments from the legal symposium are characterized by their complexity. They are reached through participant discussion and voting, some of which may lack detailed explanations. Even when a particular argument is chosen, certain statements may still be modified. Moreover, these final arguments are post-reviewed by the Criminal or Civil Divisions of the Judicial Yuan, as well as the Taiwan High Court, among others. The Judicial Yuan's research opinions and the courts' reviews may contradict the original symposium conclusions, with the reasoning and review process remaining confidential. In Task C, our accuracy assessment focuses solely on comparing the final conclusion (after review by the Judicial Yuan or Taiwan High Court) with the model's conclusion. The intermediate review process has not been considered, highlighting a limitation that requires future attention.

\subsection{Essay Questions in the Bar and Judicial Exams}
\label{sec:essay-question}

\begin{figure*}[h!]
\centering
\includegraphics[width=.85\textwidth]{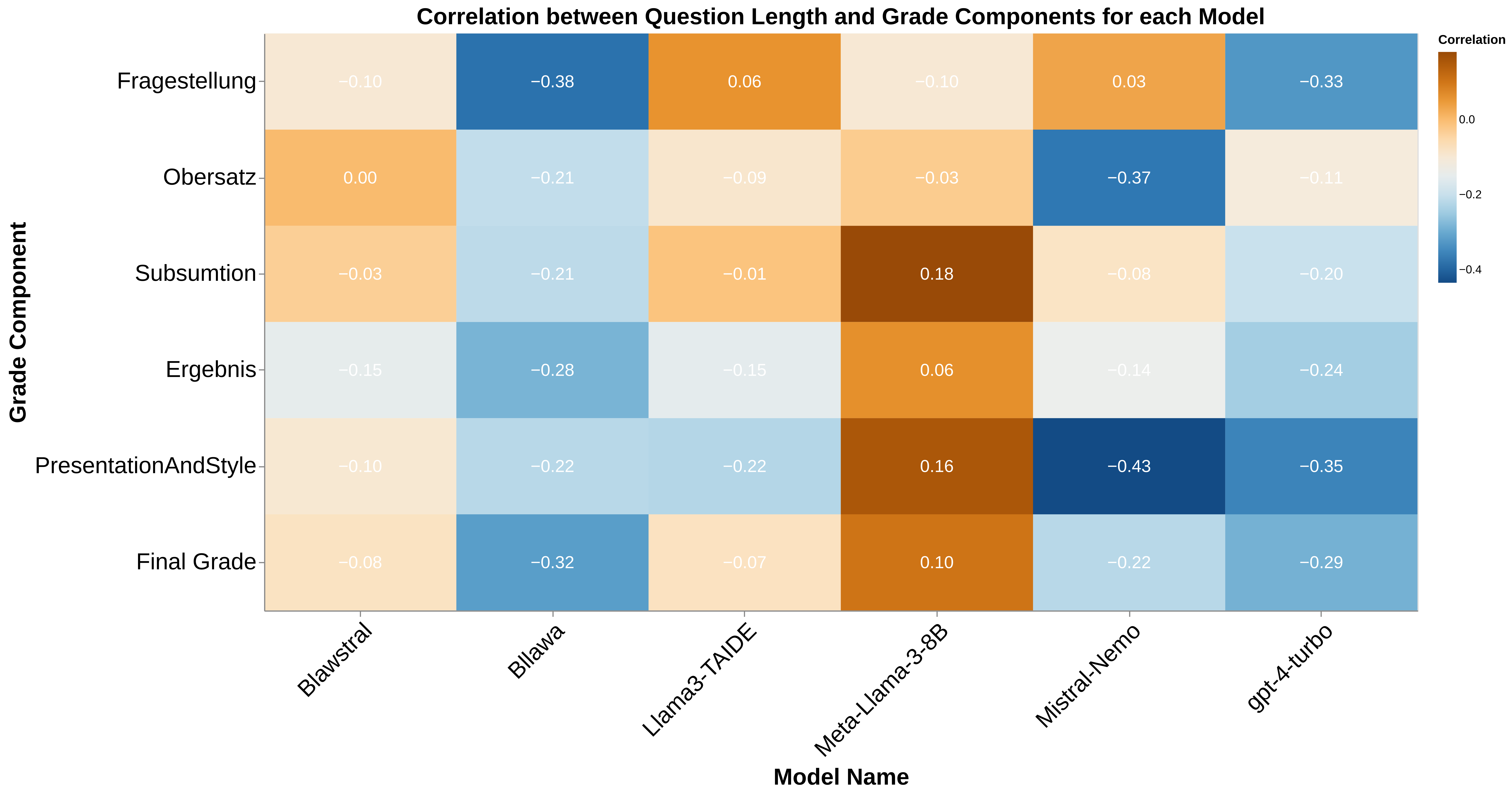}
\caption{The correlation between model performance and question length for each of the grading criteria and the final grade.}
\label{fig:question-length-vs-score-corr-1}
\end{figure*}

\begin{figure*}[h!]
\centering
\includegraphics[width=.8\textwidth]{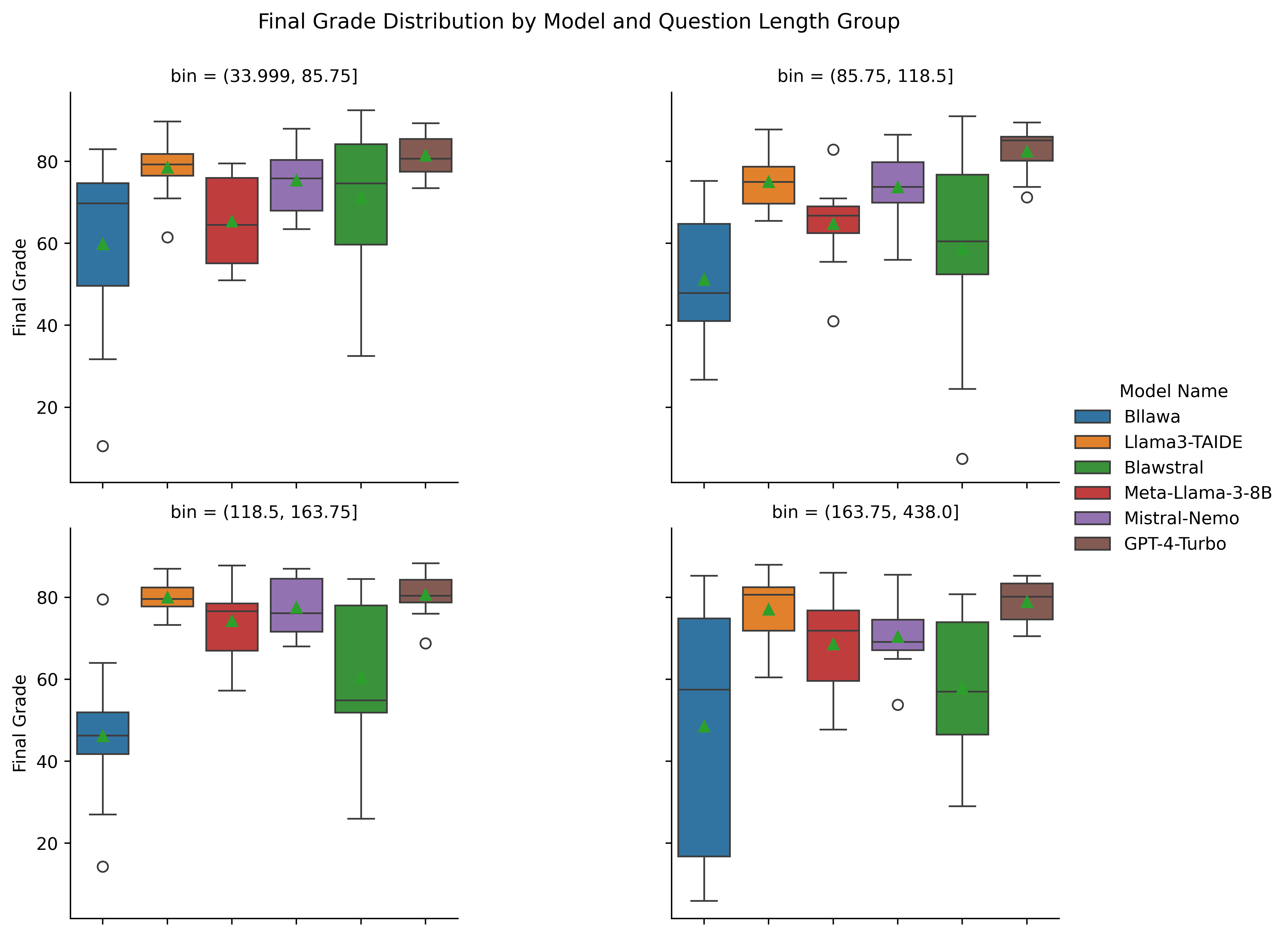}
\caption{Model performance varies as question length increases for many of the models.}
\label{fig:question-length-vs-score-corr-2}
\end{figure*}

For Task D, we use criminal law essay questions from the bar and judicial exams dataset. This kind of task has not been addressed in previous benchmarks due to the complexity involved in its evaluation. However, we believe that on the one hand, generating responses of varying lengths for this type of question may require an exploration of the impact of question length. On the other hand, this type of question represents the most challenging form of legal reasoning problems.

\modelname{GPT-4-Turbo} is prompted to segment the \textit{answers} into four chunks~\footnote{The four chunks include: Fragstellung, Obersatz, Subsumtion, and Eergebnis. This can be analogous to the IRAC reasoning steps proposed in the LegalBench benchmark: Issue, Rule, Application, and Conclusion, which is a framework American legal scholars use to execute legal reasoning.} with the German ``Juristisches Gutachten'' method, a recognized format for legal reasoning. These segmented answers are then evaluated by legal experts from law school.

We first divide the available questions into four equally sized subsets based on question length. Next, we sample ten questions from each subset, totaling forty questions. We test the following LLMs: \modelname{GPT-4-Turbo}, \modelname{Mistral-Nemo-Instruct-2407}, \modelname{Blawstral}, \modelname{Meta-Llama-3-8B-Instruct}, \modelname{Bllawa}, and \modelname{Llama3-TAIDE-LX-8B-Chat-Alpha1}.\footnote{We test these models because of their stronger performance on a pilot study.}
Given the forty essay questions, each model is required to generate answers in the format of four chunks, conforming to the Gutachten method. Finally, \modelname{GPT-4o} is used to evaluate the similarity and legal rationality between the models' answers and the golden segmented answers, with a focus on comparing each chunk individually. 

Based on the heatmap in Figure~\ref{fig:question-length-vs-score-corr-1}, it seems that the correlation between question length and the different grade components for various LLM models is relatively low, with correlations ranging between -0.29 and 0.14. The box plots in Figure~\ref{fig:question-length-vs-score-corr-2} indicate that while question length does influence the performance of certain models (like \modelname{Bllawa}), others like \modelname{GPT-4-Turbo} maintain consistent performance regardless of question length. This suggests that model architecture and training significantly influence how well a model can handle variations in question length, with more robust models being less affected by these variations. Therefore, question length might be a factor, but its impact varies depending on the specific LLM in question.

\section{Conclusion}
\label{sec:conclusion}

This paper investigated the impact of Domain-Adaptive Continual Pre-Training (DACP) on the performance of Legal LLMs in handling general and complex legal reasoning tasks. While DACP offers significant advantages by continuously integrating domain-specific knowledge, our findings indicate that its benefits are not uniformly distributed across all task types. Specifically, DACP may improve certain types of domain-specific reasoning but can simultaneously diminish the model's performance on prompt-based tasks and other generalization capabilities. These results suggest that while DACP is a valuable tool for enhancing legal reasoning in LLMs, it may not always be the optimal solution for all types of legal tasks. Future research should explore hybrid approaches that combine DACP with other methods, such as task-specific fine-tuning or meta-learning, to achieve a more balanced enhancement of both domain-specific expertise and general reasoning abilities. Additionally, further work is needed to refine evaluation benchmarks that more accurately capture the nuances of legal reasoning and the specific challenges posed by different legal tasks.

\section{Limitations}
\label{sec:limitations}

One of the limitations lies in the potential data contamination of LLMs used in our work. Since the models are pre-trained on datasets scraped from the internet, there is a risk that some of the evaluation data might overlap with the training data, which could result in the models appearing more capable than they genuinely are. 
A related issue is the ignorance of the \textit{optimal mixture ratio} between the general corpus and the legal domain-corpus~\citep{que2024dcptlawdomainspecificcontinual}.

Another limitation lies in the evaluation metrics. Traditional metrics like BLEU or ROUGE are more applicable for summarization or translation tasks, and they may fail to capture the whole context within legal reasoning tasks with more flexibility. Addressing these limitations requires the development of more robust, diverse, and human-aligned evaluation frameworks. While numerous studies adopt open-source LLMs to evaluate other LLMs' outputs, this approach introduces potential biases. This reliance raises concerns about objectivity and comprehensiveness, as we may not fully explain or account for the intricacies of how LLMs interpret and assess the data.

\section*{Acknowledgments}
We thank Angle Publishing for providing access to educational materials, the NVIDIA-NTU Artificial Intelligence Joint Research Center for providing access to the Taipei-1 supercomputer, and the NTU Research Center for Digital Law.

\bibliography{custom}
\end{CJK*}

\end{document}